\documentclass[conference]{IEEEtran}

\usepackage[letterpaper]{geometry}

\usepackage[utf8]{inputenc}
\usepackage{url}
\usepackage{amsmath}
\usepackage{amsfonts}
\usepackage{amssymb}
\usepackage{bm}
\usepackage{ltexpprt}
\usepackage{graphicx}
\graphicspath{{./figures/}}
\usepackage{caption}
\usepackage{subcaption}
\usepackage{xcolor}

\begin{document}

\bstctlcite{IEEE:BSTcontrol}

\urldef{\mails}\path|{revelle,carlotta}@cs.gmu.edu,bgelman@gmu.edu|

\title{Group-Node Attention for Community Evolution Prediction}

\author{
    \IEEEauthorblockN{Matt Revelle, Carlotta Domeniconi, Ben Gelman}
     \IEEEauthorblockA{George Mason University,\\
     Fairfax, VA 22030, USA\\
    \mails}
 }



\date{}

\maketitle







\begin{abstract} \small\baselineskip=9pt Communities in social networks evolve over time as people enter and leave the network and their activity behaviors shift. The task of predicting structural changes in communities over time is known as community evolution prediction. Existing work in this area has focused on the development of frameworks for defining events while using traditional classification methods to perform the actual prediction. We present a novel graph neural network for predicting community evolution events from structural and temporal information. The model (GNAN) includes a group-node attention component which enables support for variable-sized inputs and learned representation of groups based on member and neighbor node features. A comparative evaluation with standard baseline methods is performed and we demonstrate that our model outperforms the baselines. Additionally, we show the effects of network trends on model performance.
\end{abstract}

\section{Introduction}
Social network communities evolve over time according to the behavior of individual community members and predicting community evolution can help us in modeling the dynamics of the entire network. By focusing on community dynamics we are able to both leverage community member nodes as related data points to improve prediction performance and apply our methods to important real-world goals such as supporting study groups in massive open online courses (MOOCs) \cite{gelman2016acting, yang2015uncovering} or disrupting groups involved in criminal activity \cite{ouellet2019one}.

Existing work in community evolution prediction has focused on developing new frameworks that label events between pairs of communities -- or groups -- in consecutive network snapshots. A main goal of these frameworks is to define community evolution events in such a way that prediction of those events by standard classification models improves relative to other frameworks. Not all frameworks use the same set of evolution events, but the events from one of the first frameworks for community evolution prediction \cite{brodka2013ged} can either be mapped to events in other frameworks or are explicitly dropped. This standard set of predicted evolution events include: \emph{continuing}, \emph{dissolving}, \emph{growing}, \emph{merging}, \emph{shrinking}, and \emph{splitting}.

Given a series of network snapshots over time, the preparation of a dataset for community evolution prediction involves performing community detection on each snapshot network, finding relationships between communities in adjacent snapshots, and then using the relationships and additional network features to define community evolution events. Community evolution prediction differs from other graph prediction problems in that network communities are found and provided as input to prediction models. While previous work has utilized features derived from community structure, we can go further and use group attributes to determine the relevance of node attributes for prediction.

In this work, we propose the Group-Node Attention Network (GNAN) model, a graph neural network that uses both node and group features to predict the occurrence of one or more community evolution events in the next snapshot. A novel component of the model is group-node attention, where the node attributes of individual community members and their neighbors are used to learn a group embedding based on their relevance according to the group attributes.

To the best of our knowledge, GNAN is the first use of graph neural networks for the community evolution prediction task.

Our contributions are as follows: (1) We define a neural network architecture (GNAN), which incorporates the novel group-node attention component to learn group embeddings from individual nodes for the community evolution prediction task; (2) we perform a comparative analysis of GNAN against four leading baseline methods and show that GNAN generally outperforms them all with statistical significance; and (3) we show how model performance is affected by the network trends present in the training data.

\section{Background}
\subsection{Community Evolution Prediction}
\label{sec:comm-evol}

\begin{figure}
    \centering
    \includegraphics[width=0.45\textwidth]{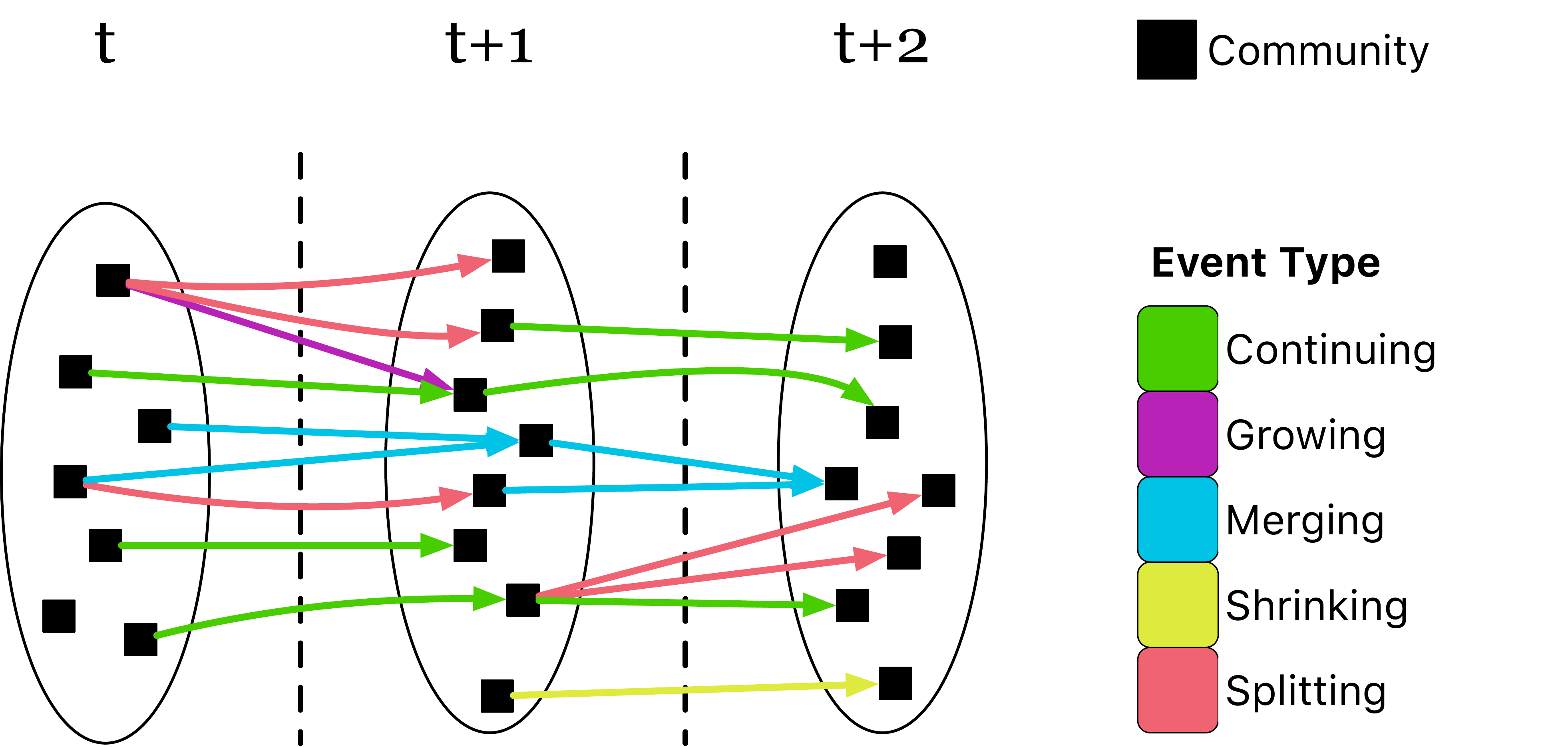}
    \caption{Community evolution events between pairs of communities in consecutive network snapshots.}
    \label{fig:sga:community-evolution}
\end{figure}

Community evolution prediction is the task of predicting future evolution events for communities over a series of network snapshots. There has been substantial work on community evolution prediction in the last 10 years that have provided frameworks for tracking evolving communities across network snapshots (as shown in Fig.~\ref{fig:sga:community-evolution}) and trained classifiers to predict future evolution events \cite{saganowski2019analysis,tajeuna2015tracking,brodka2011tracking}.

We use the Group Evolution Discovery (GED) \cite{brodka2013ged} framework to find community evolution events, but our proposed model can be used with any framework that provides community evolution event labels. Many traditional classification techniques (decision trees, SVM, etc.) have been used in previous work \cite{khafaei2019tracing,saganowski2019analysis,pavlopoulou2017predicting}. This literature often includes experiments that evaluate the contribution of engineered features \cite{pavlopoulou2017predicting,ilhan2016feature}. However, these previous works do not introduce new classification models and instead redefine the community evolution events. This acts as a form of feature engineering where the definitions of community evolution events are crafted such that prediction performance is improved. Consequently, we focus on improving prediction performance by developing our own classifier while using an existing community evolution prediction framework.

\subsection{Graph Neural Networks}
Advancements in neural network architectures are being applied to graphs, and there are several surveys \cite{lee2019attention, wu2020comprehensive, zhang2020deep} that provide an overview of the model architectures and their applications in graph classification \cite{velivckovic2017graph}, graph embedding \cite{zhang2018end}, node embedding \cite{velivckovic2017graph, nguyen2019universal}, link prediction \cite{zhang2018link}, graph generation \cite{wu2020evonet, liao2019efficient}, and heterogeneous networks \cite{zhang2019heterogeneous}. The introduction of attention to graph neural networks \cite{velivckovic2017graph} has been particularly useful for working with graph data due to supporting variable-sized, nonlinear inputs and trainable, independent aggregation weights.

Our model, the Graph-Node Attention Network (GNAN), is a graph neural network that uses \emph{group-node attention} to learn a group representation for predicting community evolution events. Group-node attention represents a community by using group-level features to attend to individual nodes associated with the community. To the best of our knowledge, there is no existing work using graph neural networks for community evolution prediction or employing group-node attention.

\section{Group-Node Attention Network (GNAN)}

\begin{table*}
  \caption{Definition of notation.}
  \begin{subtable}{0.49\textwidth}
  \centering
  \begin{tabular}{| l p{5cm} | }
    \hline
    \textbf{Symbol}                                           & \textbf{Definition} \\
    \hline
    $\{\mathcal{G}^1, \ldots, \mathcal{G}^T\}$ & Series of $T$ snapshot networks. \\
    $\mathcal{G}^t = (\mathcal{V}^t, \mathcal{E}^t)$          & The set of nodes and edges for network $\mathcal{G}^t$. \\
    $\{\mathcal{C}^1, \ldots, \mathcal{C}^T\}$ & Series of community subgraph sets from $T$ snapshot networks. \\
    $\mathcal{C}_i^t$                                         & The subgraph for group $i$ in snapshot $t$. \\
    $\mathsf{M}(\mathcal{C}_i^t),\, \mathsf{N}(\mathcal{C}_i^t)$                                & The set of member nodes and adjacent nodes for subgraph $\mathcal{C}_i$, respectively. \\
    $N_i$                                                       & Number of nodes associated with a group $i$. \\
    $E$                                                       & Number of event classes. \\
    $D_{n},\, D_{g}$                                            & Number of attributes in a single snapshot per node/vertex and group. \\
    $D_{m}$                                                   & Number of hidden dimensions. \\
    $D_q,\, D_k,\, D_v$                                           & The query, key, and value sizes used in group-node attention. \\
    \hline
  \end{tabular}
  \end{subtable}
  \begin{subtable}{0.49\textwidth}
  \centering
  \begin{tabular}{| l p{6.4cm} |}
    \hline
    \textbf{Symbol}                                           & \textbf{Definition} \\
    \hline
    $H$                                                       & Number of attention heads. \\
    $P$                                                       & Number of previous snapshots used to construct node attribute vectors. \\
    $\mathbf{x}^t_u$                                          & $1 \times D_n P$ attribute vector for node $u$ at snapshot $t$. \\
    $\mathbf{X}^t_i$                                          & $N_i\times D_n P$ attribute matrix for group $i$ in snapshot $t$. \\
    $\mathbf{Z_{X}}$, $\mathbf{z}_q$                          & $N_i \times D_m$ node representation matrix and $D_q$ group representation matrix. \\
    $\mathbf{g}^t_i$                                          & $1 \times D_g$ group attribute vector for group $i$ in snapshot $t$. \\
    $\mathbf{m}^t_i$                                          & $1 \times N_i$ group-relative position vector for all nodes in group $i$ in snapshot $t$. \\
    $\mathbf{h}_X$, $\mathbf{h}_g$                            & Hidden representation of the node and group attributes. \\
    $\mathbf{\tilde{y}}^{t+1}_i$                              & $1 \times E$ multi-label prediction vector for group $i$ at snapshot $t+1$. \\
    \hline
  \end{tabular}
  \end{subtable}
  \label{tbl:sga:notation}
\end{table*}

Given a dynamic network, we represent it as a series of network snapshots $\{\mathcal{G}^1, \ldots, \mathcal{G}^T\}$, where at each snapshot index $t$ there is a graph $\mathcal{G}^t = (\mathcal{V}^t, \mathcal{E}^t).$ For each network snapshot $\mathcal{G}^t$, there is a corresponding set of communities $\mathcal{C}^t$. The community subgraph for group $i$ in snapshot $t$ is referenced as $\mathcal{C}^t_i \in \mathcal{C}^t$. Community subgraphs are used to identify the member and neighbor nodes associated with a group.

\begin{figure*}
    \centering
    \includegraphics[width=0.75\textwidth]{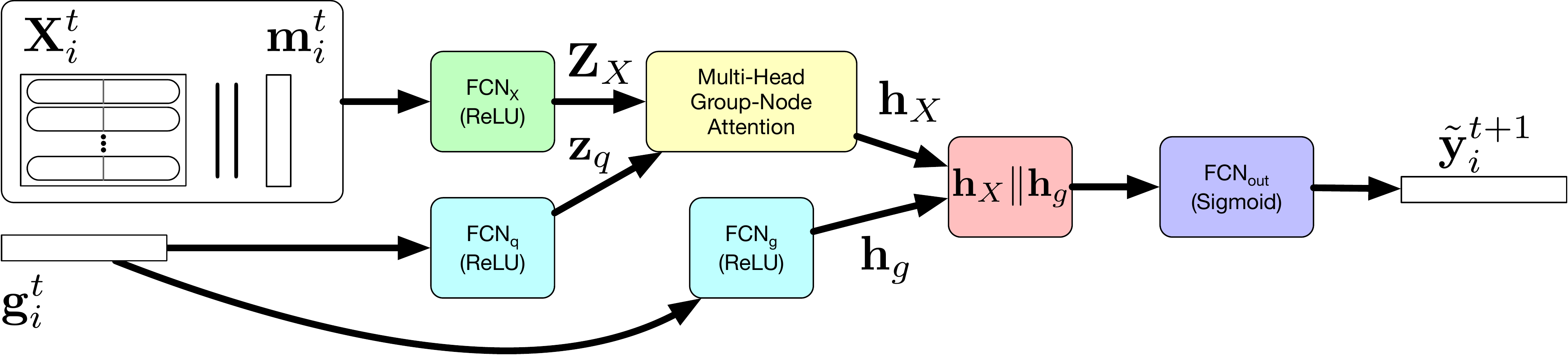}
    \caption{A diagram of the GNAN architecture showing the computation of event predictions of group $i$ at snapshot $t$. The $\|$ symbol represents concatenation.}
    \label{fig:sga:model}
\end{figure*}

The GNAN (Fig.~\ref{fig:sga:model}) accepts input values for a single group $i$ at snapshot $t$ and outputs a multi-label prediction vector for the evolution events that group $i$ will participate in with groups from the next network snapshot $t+1$. There are multiple inputs associated with each group that are accepted by the model; these are: a node attribute matrix $\mathbf{X}_i^t$, a group attribute vector $\mathbf{g}_i^t$, and a group-relative node position vector $\mathbf{m}^t_i$.

Each group $i$ in snapshot $t$ is associated with a set of nodes comprising of the group members $\mathsf{M}(\cdot)$ and group neighbors $\mathsf{N}(\cdot)$ of the community subgraph $\mathcal{C}_i^t$. The group-relative position vector $\mathbf{m}^t_i$ indicates whether a node is a group member or a neighbor. This positional information allows the model to distinguish between group members and neighbors when predicting events. The node features may include information from previous snapshots and can be derived from the network topology (e.g., in-degree) or from external data such as text documents associated with nodes. Group features capture aggregate information of group members (e.g., group size) and how group members are connected to each other or the rest of the network (e.g., edge density). GNAN can support any continuous-valued attributes.

There are three fully-connected networks (FCNs) used for input transformations that we refer to as: $\text{FCN}_X$, $\text{FCN}_q$, and $\text{FCN}_g$.

Each $\text{FCN}$ used in our implementation performs a linear transform of its input and then a non-linear activation, e.g.
\begin{equation*}
    \text{FCN}_g(\mathbf{g}^t_i) = \textsf{ReLU}(\mathbf{W}_g\mathbf{g}^t_i + \mathbf{b}_g),
\end{equation*}
where $\mathbf{W}_g \in \mathbb{R}^{D_g \times D_m}$ is a weight matrix and $\mathbf{b}_g$ is a bias vector. These transforms are used to reshape the input to match the appropriate dimensions used by the model and each has its own weight and bias parameters.

To simplify notation we will introduce several intermediate variables which correspond to the output of these transformations:
\begin{align*}
    \mathbf{z}_q &= \text{FCN}_q(\mathbf{g}^t_i), \\
    \mathbf{Z}_X &= \text{FCN}_X(\mathbf{X}^t_i \, \| \, \mathbf{m}^t_i), \\
    \mathbf{h}_g &= \text{FCN}_g(\mathbf{g}^t_i)\,,
\end{align*}
where $\|$ denotes matrix or vector concatenation.

For readability, we will drop the $i$ and $t$ annotations when referring to the intermediate variables. An individual input to the model is for a single group $i$ at a snapshot $t$. As shown in Fig.~\ref{fig:sga:model}, the node and group inputs pass through separate fully-connected networks, $\text{FCN}_X$ and $\text{FCN}_q$, which output $\mathbf{Z}_X$ and $\mathbf{z}_q$. These are used in the multi-head group-node attention component detailed in Section \ref{sec:sga:group-node-att}. The group attribute vector $\mathbf{g}^t_i$ is also used to compute a hidden representation $\mathbf{h}_g$ for the group $i$ using only the group features.

Finally, the output of group-node attention $\textsc{GNAtt}(\cdot)$ is concatenated with $\mathbf{h}_g$ and passed through an output layer
\begin{align*}
    \mathbf{h}_X &= \textsc{GNAtt}(\mathbf{Z}_X, \mathbf{z}_q)\,, \\
    \mathbf{\tilde{y}} &= \text{FCN}_{\textsf{out}}(\mathbf{h}_X \, \| \, \mathbf{h}_g)\,,
\end{align*}
where $\text{FCN}_{\textsf{out}}$ uses a sigmoid activation function to scale the predicted class label probabilities, $\mathbf{\tilde{y}} \in \mathbb{R}^{E}$; and $E$ is the number of community evolution event classes.

The details of constructing the node attribute matrix $\mathbf{X}_i^t$ and the group-node attention component are provided in Sections \ref{sec:sga:mixing} and \ref{sec:sga:group-node-att}.

\subsection{Spatial and Temporal Mixing}
\label{sec:sga:mixing}

A major intuition behind our model is to incorporate changes over time. For a single node $u$ which exists in both snapshots $\mathcal{G}^{t-1}$ and $\mathcal{G}^{t}$ we can form a vector $\mathbf{x}^t_u$ that is a concatenation of the attributes of node $u$ at snapshots $t-1$ and $t$. More generally, given consecutive network snapshots $\{\mathcal{G}^{t-1}, \mathcal{G}^t\}$ and sets of communities found in each snapshot $\{\mathcal{C}^{t-1}, \mathcal{C}^t\}$, we can form the input node attributes matrix $\mathbf{X}^t_i$ for group $i$ at snapshot $t$. This input matrix $\mathbf{X}^t_i$ is constructed through row concatenation of all node feature vectors $\mathbf{x}^t_u$ for each node $u$ that is a member or neighbor of group $i$: $\{u \in \mathsf{M}(\mathcal{C}^t_i) \, \cup \, \mathsf{N}(\mathcal{C}^t_i)\}$. An additional dimension, represented as vector $\mathbf{m}^t_i$ in Fig.~\ref{fig:sga:model}, is used to specify whether the node is a member or neighbor (one-hop neighbor of the group) and adds spatial position information to the node attribute vectors.

\subsection{Group-Node Attention}
\label{sec:sga:group-node-att}
\begin{figure*}[ht]
    \centering
    \includegraphics[width=0.9\textwidth]{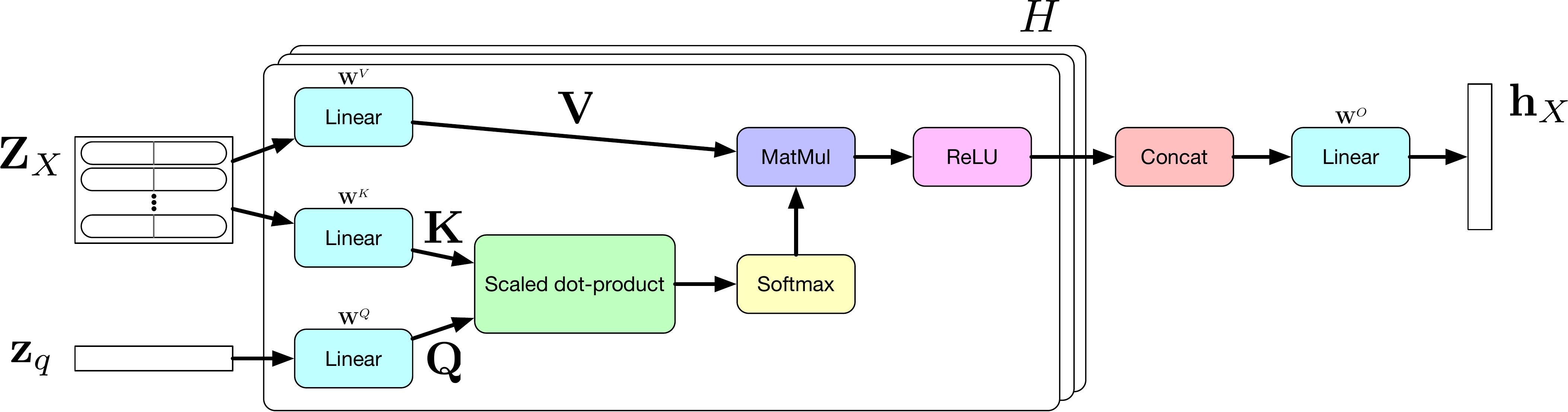}
    \caption{A diagram of the multi-head group-node attention layer used in the model with $H$ attention heads.}
    \label{fig:sga:group-att}
\end{figure*}

In our model, we use group-node attention rather than self-attention. That is, we treat the group features as the query and the node features of member and neighbor nodes as the keys and values, resulting in a hidden representation that uses group features to attend to nodes. The output of the attention layer is then an aggregation of the context vectors associated with group $i$ into a single hidden representation $\mathbf{h}_X$.

In order to generate $\mathbf{h}_X$, we use $\mathbf{Z}_X$, the transformed result of the spatial and temporal mixing of the input nodes,  along with $\mathbf{z}_q$, the transformed group features, to construct query, key, and value matrices ($\mathbf{Q}, \mathbf{K}, \mathbf{V}$) using learned weight matrices $\mathbf{W}^Q, \mathbf{W}^K, \mathbf{W}^V$ as in \cite{vaswani2017attention}. Fig.~\ref{fig:sga:group-att} provides a diagram of the multi-head group-node attention portion of the model where we see the separate linear layers associated with each of the weight matrices. The $\mathbf{z}_q$ vector contributes to the query matrix $\mathbf{Q}$ and $\mathbf{Z}_X$ contributes to the key and value matrices $\mathbf{K}$ and $\mathbf{V}$.

The group-node attention layer uses multi-head attention in order to potentially learn complementary group representations. The output of each head is provided by $\textsc{Head}(\cdot)$ defined as
\begin{equation*}
    \textsc{Head}(\mathbf{Z}_X, \mathbf{z}_q) = \bm{\alpha} \mathbf{V},
\end{equation*}
where $\bm{\alpha}$ is a attention coefficient vector and $\mathbf{V}$ is a value matrix.

We use scaled dot-product attention where the attention coefficients are
\begin{equation*}
    \bm{\alpha} = \textsf{softmax}(\frac{\mathbf{Q} \mathbf{K}^{\mathsf{T}}}{\sqrt{D_m}}).
\end{equation*}
The query $\mathbf{Q}$, key $\mathbf{K}$, and value $\mathbf{V}$ matrices are defined as
\begin{align*}
    \mathbf{Q} &= \mathbf{z}_q \mathbf{W}^Q, \\
    \mathbf{K} &= \mathbf{Z}_X \mathbf{W}^K, \\
    \mathbf{V} &= \mathbf{Z}_X \mathbf{W}^V
\end{align*}
where $\mathbf{W}^Q$, $\mathbf{W}^K$, and $\mathbf{W}^V$ are learned weight parameters; and $\mathbf{Q} \in \mathbb{R}^{1 \times D_k}$, $\mathbf{K} \in \mathbb{R}^{N_i \times D_k}$, and $\mathbf{V} \in \mathbb{R}^{N_i \times D_m}$. Fully expanded, the output of each head is then:
\begin{equation*}
    \textsc{Head}(\mathbf{Z}_X, \mathbf{z}_q) = \textsf{softmax}(\frac{\mathbf{z}_q \mathbf{W}^Q(\mathbf{Z}_X \mathbf{W}^K)^{\mathsf{T}}}{\sqrt{D_m}})\mathbf{Z}_X\mathbf{W}^V.
\end{equation*}

The head outputs are passed through a $\textsf{ReLU}$ activation and then all are concatenated ($\|$) together and sent through a final transformation,
\begin{equation*}
    \textsc{GNAtt}(\mathbf{Z}_X, \mathbf{z}_q) = (\|^H_{h} \: \textsf{ReLU}(\textsc{Head}_h(\mathbf{Z}_X, \mathbf{z}_q))) \mathbf{W}^O,
\end{equation*}
where $\mathbf{W}^O \in \mathbb{R}^{H D_v \times D_m}$ is a weight matrix parameter that mixes the results of the group heads.

Finally, the output of the multi-head group-node attention component is provided as $\mathbf{h}_X = \textsc{GNAtt}(\mathbf{Z}_X, \mathbf{z}_q)$ to the rest of the network.

\section{Experiments}

\subsection{Community Detection and Tracking}
Predicting events over a series of network snapshots where communities can be simultaneously involved in multiple evolution events requires both a method for detecting communities and tracking them across snapshots. As outlined in Section \ref{sec:comm-evol}, there are many existing community tracking frameworks. Our model does not depend on any specific framework and we selected GED \cite{brodka2013ged} for our experiments as it is well-established in the literature and supports overlapping communities. GED uses two parameters -- $\alpha$ and $\beta$ -- for labelling events; we use values of 0.5 for both parameters. We use the clique percolation method (CPM) \cite{palla2005uncovering} on clique graphs as described in \cite{evans2010clique} to define the communities for each network snapshot. Our implementation of CPM constructs a clique graph and then merges cliques that share a majority of members. CPM supports overlapping communities and uses cliques as primitives for constructing communities. This matches the intuition of our model and expected structure of social interaction networks where communities are dense, overlapping graph regions \cite{palla2005uncovering}.

\subsection{Datasets}

We use two datasets of timestamped, directed interactions to construct network snapshots. The first dataset is a collection of Facebook wall posts \cite{viswanath2009evolution} available from KONECT\footnote{\url{http://konect.cc/networks/facebook-wosn-wall/}}. In Facebook, users may post on each other's walls and these posts are typically comments, photos, and web links. Each of these posts is recorded as an interaction with a source user (the post author), a destination user (the owner of the wall), and a timestamp. The second dataset is a collection of Scratch project comments \cite{resnick-scratch} extracted from a general Scratch dataset available from the Harvard Dataverse\footnote{\url{https://dataverse.harvard.edu/dataset.xhtml?persistentId=doi:10.7910/DVN/KFT8EZ}}. Scratch is an online social network and web application for writing and sharing software projects. Users can comment on each other's projects; each project comment is recorded as an interaction between two users. Most communities (95\%) from the Facebook dataset contain six or fewer members and similarly for Scratch most communities contain eight or fewer members.

To construct network snapshots, we adopt a methodology used by \cite{miritello2013limited, navarro2017temporal, revelle2017temporal} to determine a fixed temporal window length that prevents artificial accumulation of edges. Edges are included in a snapshot if the node pair has interacted within the temporal window captured by the snapshot. The Facebook dataset extends from July 2006 until April 2009 and each network snapshot includes activity for a month. The Scratch dataset includes data from May 2010 until May 2011 and each snapshot includes two weeks worth of activity.

\begin{figure}
    \centering
    \includegraphics[width=0.85\linewidth]{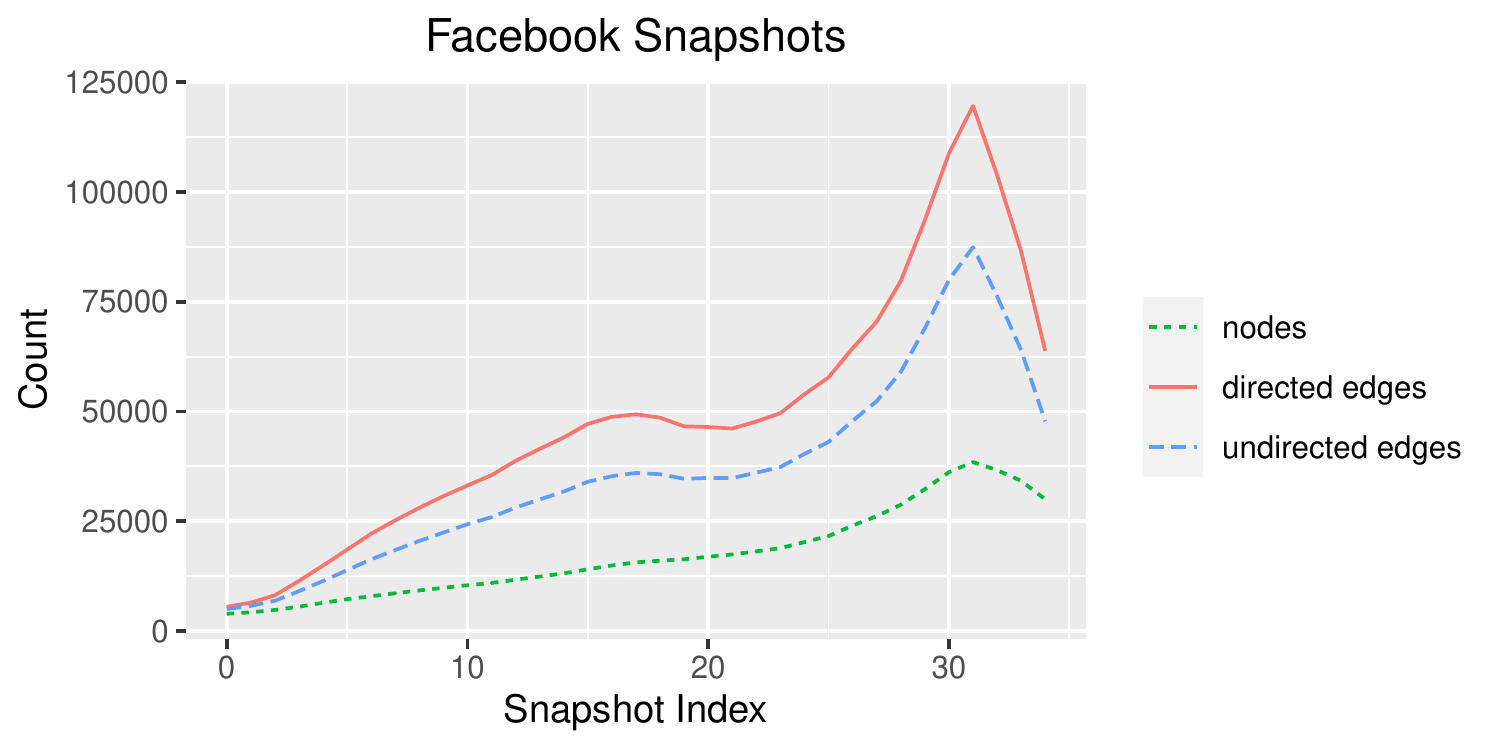}
    \caption{Facebook snapshot network size statistics.}
    \label{fig:sga:fb-net-stats}
\end{figure}

\begin{figure}
    \centering
    \includegraphics[width=0.85\linewidth]{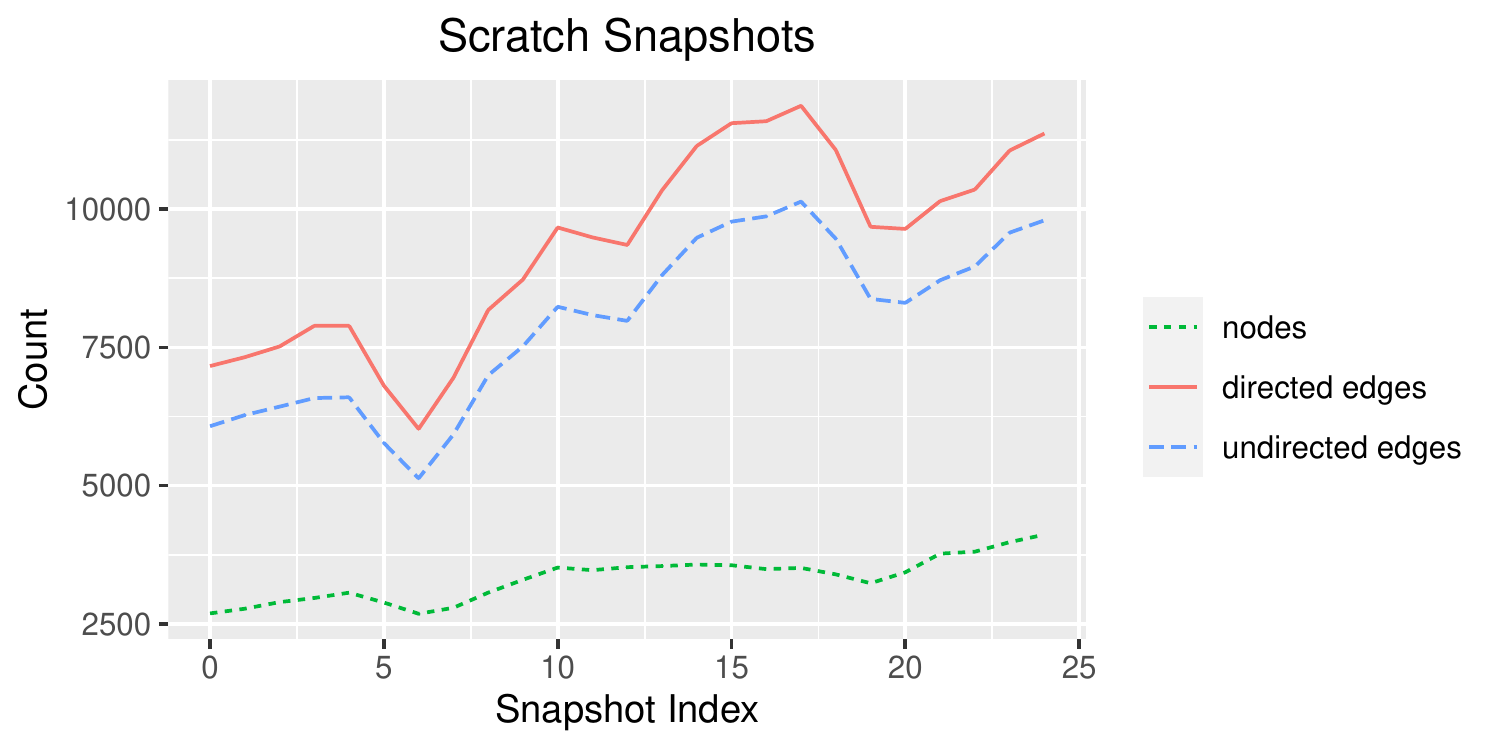}
    \caption{Scratch snapshot network size statistics.}
    \label{fig:sga:scratch-net-stats}
\end{figure}

Figs.~\ref{fig:sga:fb-net-stats} and \ref{fig:sga:scratch-net-stats} show the number of nodes and edges for both the Facebook and Scratch network snapshots. We note that \emph{continuing} and \emph{dissolving} events are the most frequent in the Facebook dataset, while \emph{merging} and \emph{splitting} are the most common in the Scratch dataset. This may be due to the nature of the networks---Facebook is primarily used to connect with people already known by a user, but Scratch encourages creating new relationships through project collaboration.

\begin{table}
  \caption{Node and group features used in experiments.}
  \begin{subtable}{0.45\textwidth}
  \centering
  \caption{Node features}
  \begin{tabular}{r l p{5cm}}
    & \textbf{Name} & \textbf{Description} \\
    1 &   \emph{In-degree} & Count of incoming edges at snapshot $t$ \\
    2 &   \emph{Out-degree} & Count of outgoing edges at snapshot $t$ \\
    3 &   \emph{Previous in-degree} & Count of incoming edges at snapshot $t-1$ \\
    4 &   \emph{Previous out-degree} & Count of outgoing edges at snapshot $t-1$ \\
  \end{tabular}
  \label{tbl:sga:node-features}
  \end{subtable}
  \vspace{1em}
  \par
  \begin{subtable}{0.45\textwidth}
  \centering
  \caption{Group features}
  \begin{tabular}{r l p{5cm}}
    & \textbf{Name} & \textbf{Description} \\
    1 &  \emph{Density} & The ratio of actual edges to potential edges among group members \\
    2 &  \emph{Group affinity} &  The ratio of edges between group members over all edges of all group members \\
    3 &  \emph{Size} & Number of member nodes \\
    4 &   \emph{Event counts} & Counts of the incoming event types from the previous snapshot\\
  \end{tabular}
  \label{tbl:sga:group-features}
  \end{subtable}
  \label{tbl:sga:features}
\end{table}

\subsection{Methodology}
\label{sec:sga:methodology}
For comparison to GNAN, we chose four baseline methods due to their diversity and frequent use for community evolution prediction. The baseline methods are CART decision trees, logistic regression, multilayer perceptron (MLP), and SVM with an RBF kernel. The implementations from \cite{scikit-learn} are used for CART, logistic regression, and SVM. Those implementations support a class weights parameter which is configured to \emph{balanced} to adjust for the class imbalance in the dataset. None of these three baselines directly support multi-label classification and an instance of the classifier was trained for each community evolution event label. We performed a sweep over the regularization parameter (\emph{C}) for SVM using the values: 0.01, 0.1, 1, 10, and 100. The GNAN and MLP models are both implemented with \cite{pytorch}, use a model size of 16 dimensions for embedding layers, and were optimized using AdamW with a learning rate of 0.001 and weight decay of 0.01. The MLP models have three layers. The binary cross-entropy loss function was used and training stops after five consecutive epochs without a decrease in validation loss. We use a basic set of network features, listed in Table \ref{tbl:sga:features}, which capture connectivity and structure. Event counts for incoming events from the previous snapshot are also included as a group feature. Since none of the baselines support variable-sized input, the features for group member and neighbor nodes were provided as two separate vectors---one for group members and one for group neighbors---containing the mean average of the feature values. Those two vectors were concatenated with the group features to form a single vector input.

We use a holdout method with random splits for evaluation. We perform 30 random splits of the consecutive network snapshots into training, validation, and test sets. All network snapshots before the split are used for training and validation, the remaining future network snapshots can be used for testing. The random splits are selected from the interval $[5, T]$ where $T$ is the final snapshot. This guarantees a minimum number of training examples are made available to the models. In order to address the non-determinism and sensitivity to parameter values for some of the models included in this experiment, we train five instances of each model for each random split. We select one of the five models for each split based on the macro-averaged AUC to use for evaluation with the test data. We only use the groups from the snapshot after the split for evaluation.

\subsection{Comparative Results}
\label{sec:comparative}

\begin{table*}
  \caption{Mean AUC scores for events in the Facebook and Scratch datasets. Highest values are in bold. The $^{\bullet}$/$^\circ$ annotation indicates whether GNAN is statistically superior/inferior to the other method. A two-sided Wilcoxon signed-rank test was used at 95\% significance level.}
\centering
  \begin{subtable}{0.75\textwidth}
  \centering
  \caption{Facebook results}
  \begin{tabular}{l | c c c c c c | c}
    Method & Cont. & Dis. & Grow & Merge & Shrink & Split & Macro Avg. \\
    \hline
    CART & $0.526^{\bullet}$ & $0.515^{\bullet}$ & $0.508^{\bullet}$ & $0.532^{\bullet}$ & $0.570^{\bullet}$ & $0.584^{\bullet}$ & $0.539^{\bullet}$ \\
    Log. Reg. & $0.581^{\bullet}$ & $0.590^{\bullet}$ & $0.552^{\bullet}$ & $0.695^{\bullet}$ & $0.923^{\bullet}$ & $0.892^{\bullet}$ & $0.706^{\bullet}$ \\
    MLP & $0.620$ & $0.596^{\bullet}$ & $0.571^{\bullet}$ & $0.718$ & $\textbf{0.939}$ & $0.962^{\bullet}$ & $0.734^{\bullet}$ \\
    SVM & $0.585^{\bullet}$ & $0.590^{\bullet}$ & $0.556^{\bullet}$ & $0.693^{\bullet}$ & $0.916^{\bullet}$ & $0.884^{\bullet}$ & $0.704^{\bullet}$ \\
    GNAN & \textbf{0.636} & $\textbf{0.617}$ & $\textbf{0.603}$ & \textbf{0.757} & \textbf{0.939} & \textbf{0.966} & $\textbf{0.753}$ \\
  \end{tabular}
  \label{tbl:sga:mean-auc-fb}
  \end{subtable}
  \vspace{1em}
  \par
  \begin{subtable}{0.75\textwidth}
  \centering
  \caption{Scratch results}
  \begin{tabular}{l | c c c c c c | c}
    Method & Cont. & Dis. & Grow & Merge & Shrink & Split & Macro Avg. \\
    \hline
    CART & $0.533^{\bullet}$ & $0.590^{\bullet}$ & $0.517^{\bullet}$ & $0.642^{\bullet}$ & $0.570^{\bullet}$ & $0.670^{\bullet}$ & $0.586^{\bullet}$ \\
    Log. Reg. & $0.588^{\bullet}$ & 0.784 & $0.584^{\bullet}$ & 0.800 & $0.681^{\bullet}$ & $0.828^{\bullet}$ & $0.710^{\bullet}$ \\
    MLP & 0.621 & 0.814 & 0.610 & 0.819 & 0.778 & \textbf{0.908} & 0.756 \\
    SVM & $0.588^{\bullet}$ & 0.784 & $0.569^{\bullet}$ & $0.749^{\bullet}$ & $0.751^{\bullet}$ & $0.824^{\bullet}$ & $0.710^{\bullet}$ \\
    GNAN & \textbf{0.631} & \textbf{0.827} & \textbf{0.636} &
    \textbf{0.845} & \textbf{0.782} & 0.907 &  \textbf{0.769} \\
  \end{tabular}
  \label{tbl:sga:mean-auc-scratch}
  \end{subtable}
  \label{tbl:sga:mean-auc}
\end{table*}

The results of GNAN against the baselines are shown in Table \ref{tbl:sga:mean-auc}. The mean AUC scores are used for the evaluation as they capture the overall comparative performance. A two-sided Wilcoxon signed-rank test is used at a 95\% significance level to compare scores. We find that GNAN outperforms all baselines on both datasets with the exception of the shrinking event in Facebook and the splitting event in Scratch. GNAN achieves a statistically significant higher macro-average mean AUC over all baselines on Facebook, and all baselines but the MLP for Scratch.

In the Facebook snapshot series, we see that all methods other than CART perform well for predicting shrinking and splitting events. This is likely due to group size and previous snapshot event counts being good indicators of shrinking and splitting. We notice the same is true for the splitting event in the Scratch dataset, but prediction of the shrinking event seems to be more challenging.

The Scratch networks were constructed from social interactions and we required there be at least four interactions between a pair of nodes before including the edge in the networks. This was done in order to improve the performance of community detection with CPM; however, we expect that including that missing structural information would further improve the relative performance of GNAN over the baselines for the Scratch network snapshots.

\subsection{Temporal Effects}

In addition to the comparative experiment, we evaluate the performance of GNAN at particular snapshots and differing amounts of training history. These experiments were conducted on both the Facebook and Scratch snapshots, but only figures for Facebook are included for brevity.

\begin{figure}
    \centering
    \includegraphics[width=1.0\linewidth]{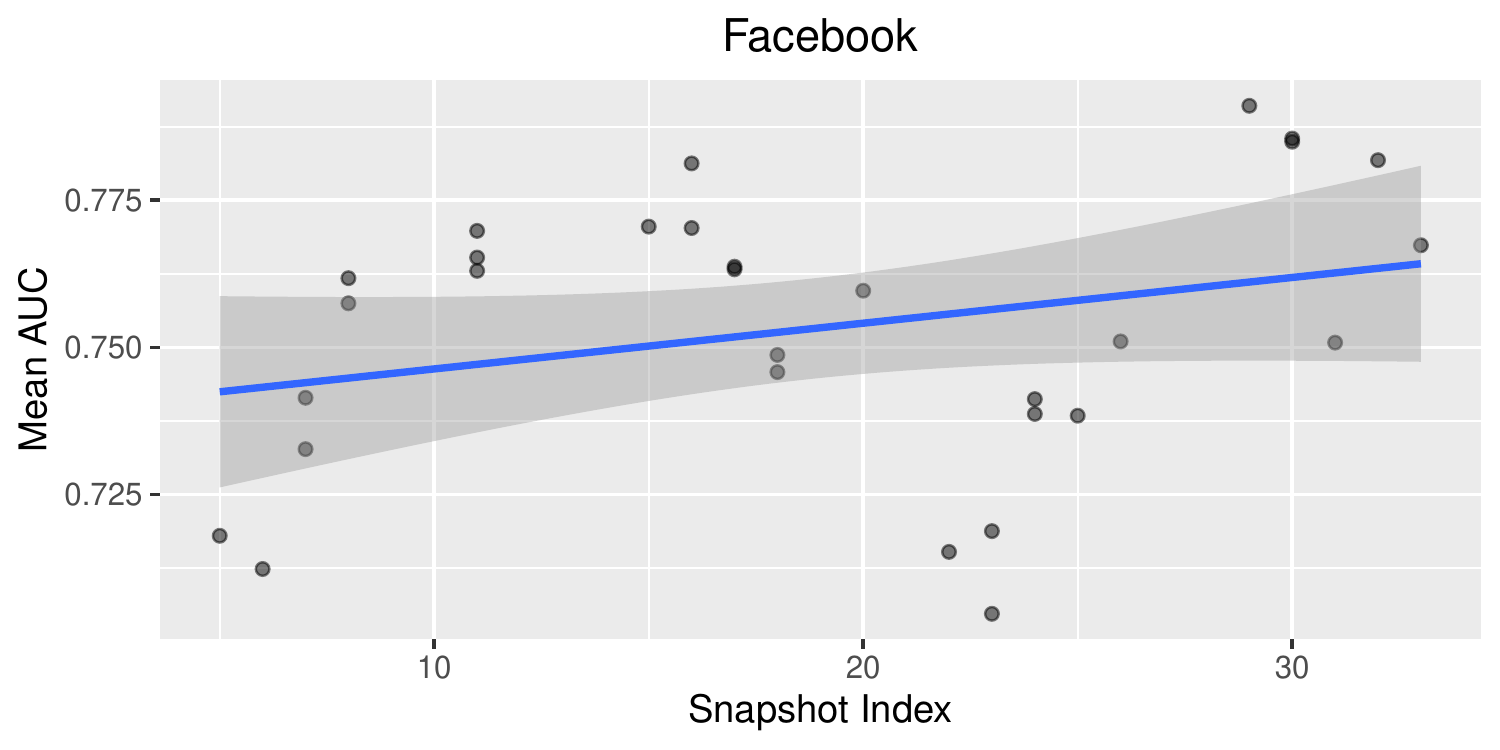}
\caption{The mean AUC scores for GNAN on the Facebook network snapshots.}
\label{fig:sga:snapshot-scores}
\end{figure}

In Fig.~\ref{fig:sga:snapshot-scores}, we see the mean AUC scores for the GNAN model instances trained for the comparative analysis (Section \ref{sec:comparative}) on the Facebook network snapshots. If we consider the changes in network activity over the snapshots in Fig.~\ref{fig:sga:fb-net-stats}, we can see that GNAN performance appears to correlate with network growth. Using the number of undirected edges as an indicator of network activity, we calculate Spearman's rank correlation coefficient and find that the mean AUC is slightly correlated with the activity for the Facebook snapshots, $\rho = .2956,\; p = 0.1$, and for the Scratch snapshots, $\rho = .3501,\; p = .06$.

A decrease in prediction performance during periods of reduced network activity may indicate that the model is missing additional information useful for predicting community evolution events. Notably, events external to the network -- such as start/stop of academic semesters or holidays -- may impact node behavior.

\begin{figure}
\begin{subfigure}{.5\textwidth}
    \centering
    \includegraphics[width=1.0\linewidth]{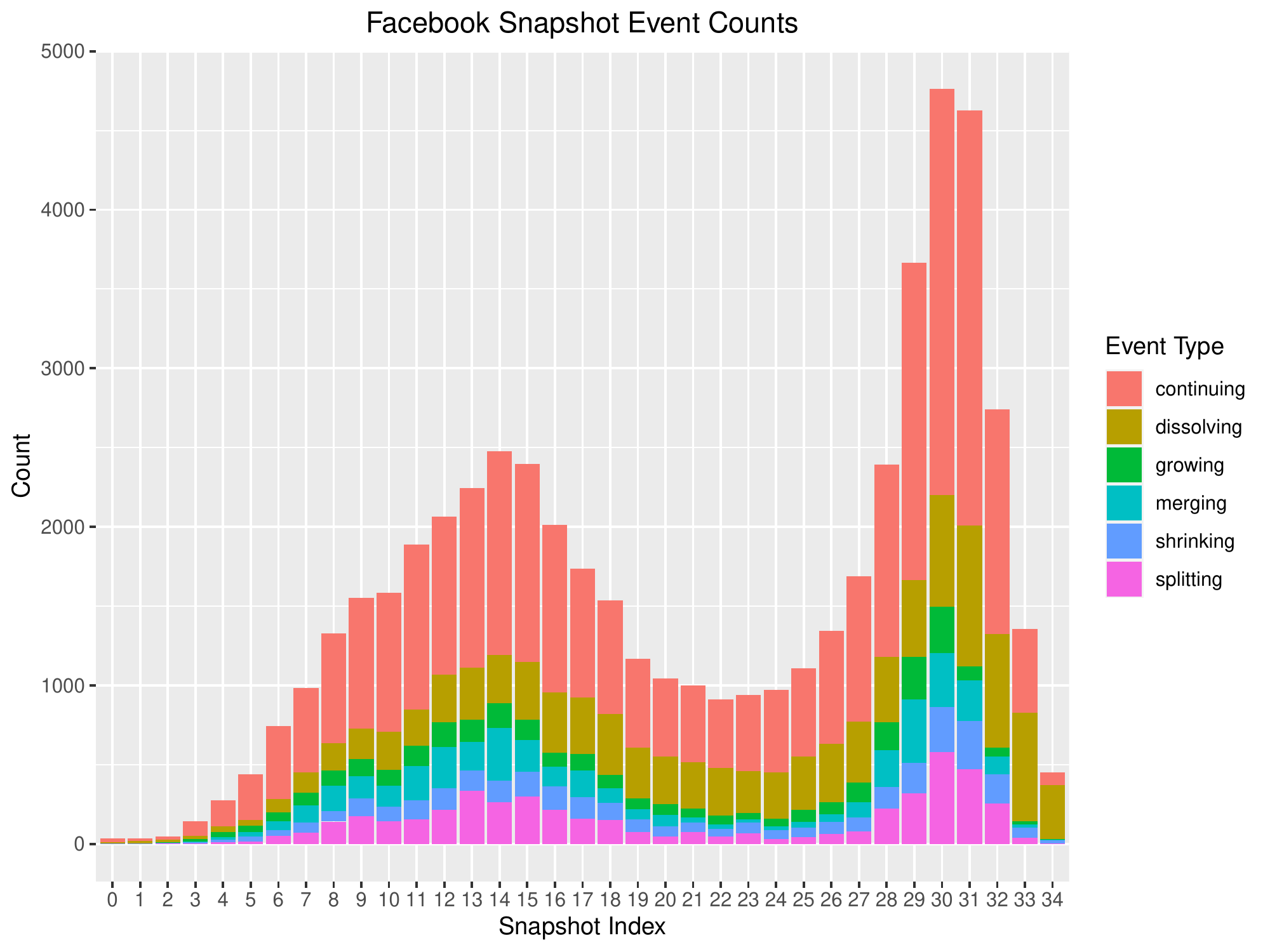}
    \caption{Facebook event counts}
    \label{fig:sga:snapshots-events:fb-counts}
\end{subfigure}
\begin{subfigure}{.5\textwidth}
    \centering
    \includegraphics[width=1.0\linewidth]{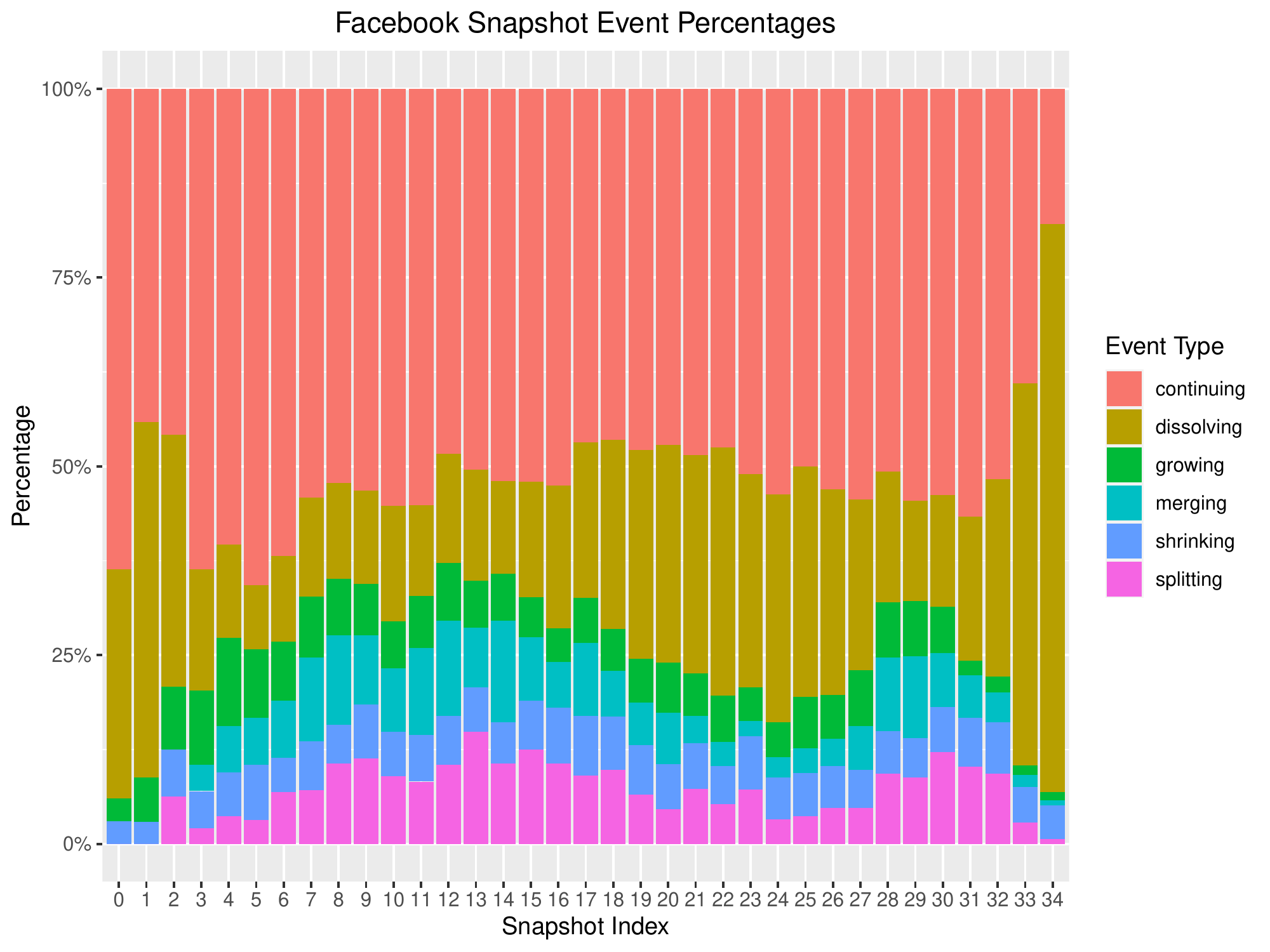}
    \caption{Facebook event percentages}
\end{subfigure}
\caption{The event counts and percentages over network snapshots for the Facebook dataset.}
\label{fig:sga:snapshot-events-fb}
\end{figure}

Additionally, we see the evolution event counts in Facebook (Fig.~\ref{fig:sga:snapshot-events-fb}) are positively correlated with network activity (Fig.~\ref{fig:sga:fb-net-stats}; and we observed the same for the Scratch snapshots. The event percentage plot for the Facebook snapshots in Fig.~\ref{fig:sga:snapshot-events-fb} reveals that the distribution of events changes over time. Though not shown here, we observed similar changes in the Scratch snapshots. In the Facebook snapshots, the proportion of dissolving events is negatively correlated with increased network activity, while continuing, growing, merging, and splitting events all appear to be positively correlated with increased network activity.

While all previous snapshots were used in training GNAN and the baseline models for the comparative evaluation in Section \ref{sec:comparative}, additional GNAN model instances were trained with incrementally larger snapshot intervals to determine the performance impact of expanding training data by incrementally adding older network snapshots. Snapshots for evaluation were selected by starting at the final snapshot of each dataset and adding every fifth snapshot index. For each evaluation snapshot index, training data was provided in increasingly larger intervals with a stride of five. When fewer than five snapshots remain then all are added to the final interval of training snapshots. For example, for the evaluation snapshot index of 14, there are three training intervals: $[9, 13]$, $[4, 13]$, $[1, 13]$. Five instances of GNAN are trained for each of the three intervals of training data and then validation is used to select the best model from each training interval for evaluation.

\begin{figure}
    \centering
    \includegraphics[width=1.0\linewidth]{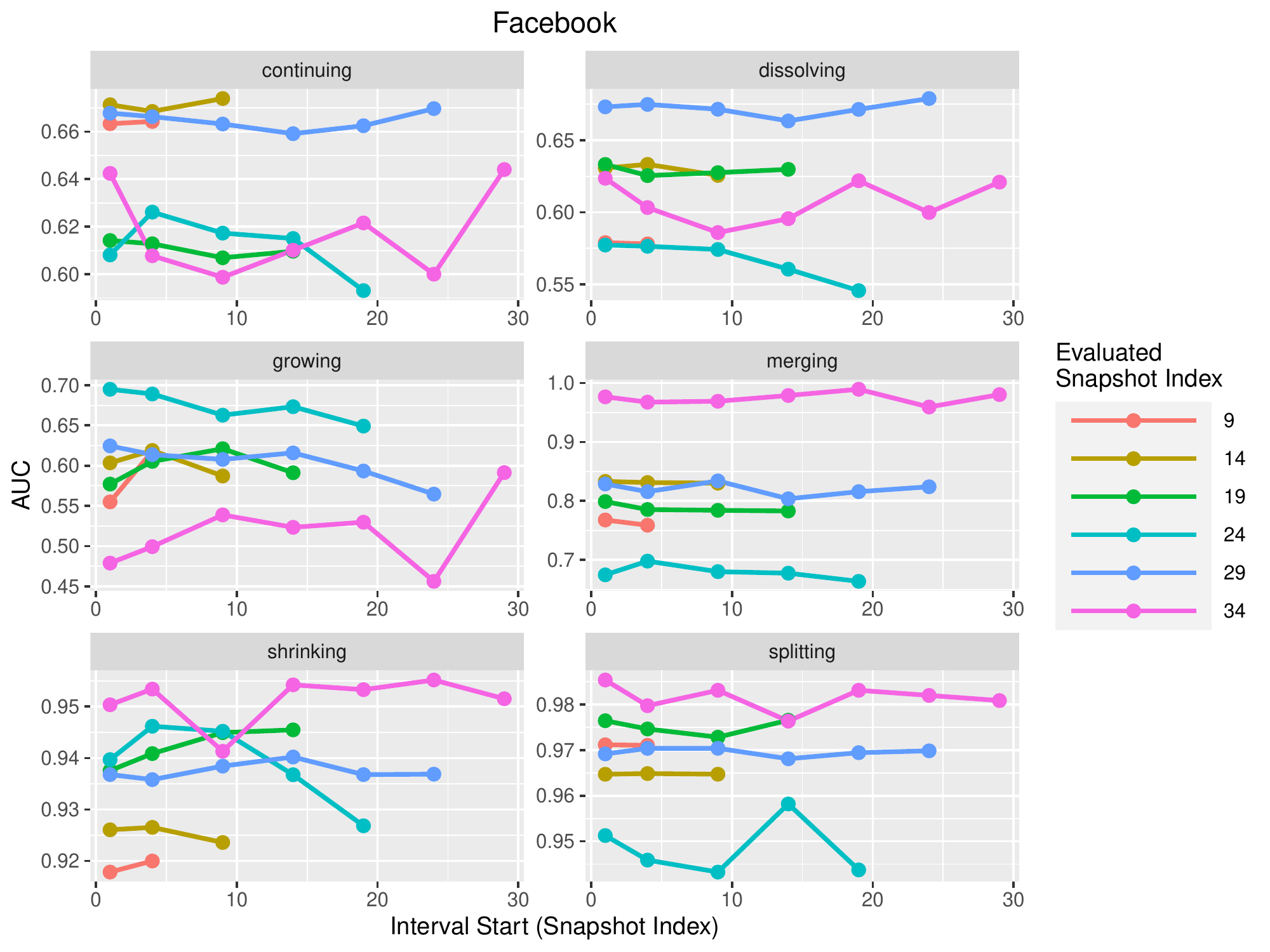}
    \caption{GNAN model performance as earlier training data is introduced for the Facebook dataset.}
    \label{fig:sga:temporal_training_scores_fb}
\end{figure}

The results for Facebook (Fig.~\ref{fig:sga:temporal_training_scores_fb}) and Scratch show that while generally more training data improves model performance, including data from earlier snapshots can negatively impact prediction performance for certain evolution events. The network activity (Figs.~\ref{fig:sga:fb-net-stats} and \ref{fig:sga:scratch-net-stats}) of Facebook and Scratch and the distributions of evolution events (Fig.~\ref{fig:sga:snapshot-events-fb}) change over time and this can affect model training.

For Facebook, the prediction performance of GNAN on shrinking and splitting events appears to be consistent across all evaluation snapshots. While not as tightly grouped as the Facebook AUC scores, the shrinking and splitting events also have the lowest variance in score across snapshots for the Scratch dataset. The prediction performance for all other events in both datasets appear to be more dependent on the evaluation snapshot used.

Consider the prediction of growing events for the Facebook evaluation snapshot at index 34. We see that using only snapshots 29-33 improves performance compared to using snapshots 24-33. However, adding the five next earlier snapshots such that all snapshots 19-33 are used increases performance again. According to Fig.~\ref{fig:sga:snapshots-events:fb-counts}, there are growing training examples gained by including all of the earlier snapshots. From Fig.~\ref{fig:sga:fb-net-stats}, we see that while the number of edges decline for a period over snapshots 29-33 and snapshots 19-23, there is only a growth of the number of edges in the snapshots 24-29. It appears more training examples for the growing event only improve prediction performance when those examples are taken from snapshots with network activity similar to that of the evaluation snapshot. This relationship between network activity in the training snapshots and the evaluation snapshot suggests that model performance may be improved by selecting training snapshots that capture similar network trends as the evaluation snapshot.

\section{Conclusion}

We introduced a graph neural network with group-node attention (GNAN) for community evolution prediction. GNAN is able to incorporate both spatial and temporal information of individual member and neighbor nodes by way of group-node attention. The model is capable of learning a task-specific group representation for community evolution prediction and outperforms the typical baselines used for predicting community evolution events.

\bibliographystyle{IEEEtran}
\bibliography{comm-dyn}

\end{document}